\documentclass[letterpaper]{article} % DO NOT CHANGE THIS
\usepackage{aaai2026}  % DO NOT CHANGE THIS
\usepackage{times}  % DO NOT CHANGE THIS
\usepackage{helvet}  % DO NOT CHANGE THIS
\usepackage{courier}  % DO NOT CHANGE THIS
\usepackage[hyphens]{url}  % DO NOT CHANGE THIS
\usepackage{graphicx} % DO NOT CHANGE THIS
\usepackage{natbib}  % DO NOT CHANGE THIS AND DO NOT ADD ANY OPTIONS TO IT
\usepackage{caption} % DO NOT CHANGE THIS AND DO NOT ADD ANY OPTIONS TO IT
\usepackage{algorithm}
\usepackage{algorithmic}

\usepackage{multirow}
\usepackage{array}
\usepackage{comment}
\usepackage{float}
\usepackage{pifont}
\usepackage{xspace}
\usepackage{booktabs}
\usepackage{amsmath}
\usepackage{xcolor}

\newcommand{\systemnm}{StoryTeller\xspace}
\newcommand{\datasetnm}{MovieStory101\xspace}
\newcommand{\qanm}{StoryQA\xspace}

\definecolor{cgreen}{RGB}{6,176,80}
\definecolor{cred}{RGB}{192,0,0}

\usepackage{newfloat}
\usepackage{listings}
\DeclareCaptionStyle{ruled}{labelfont=normalfont,labelsep=colon,strut=off} % DO NOT CHANGE THIS
\floatstyle{ruled}
\newfloat{listing}{tb}{lst}{}
\floatname{listing}{Listing}
\title{\systemnm: Improving Long Video Description through Global Audio-Visual Character Identification}
\author{
    Yichen He\textsuperscript{\rm 1}\equalcontrib,
    Yuan Lin\textsuperscript{\rm 1}\equalcontrib,
    Jianchao Wu\textsuperscript{\rm 1}\equalcontrib,
    Hanchong Zhang\textsuperscript{\rm 2},
    Yuchen Zhang\textsuperscript{\rm 1}\thanks{Corresponding author.},
    Ruicheng Le\textsuperscript{\rm 3}
}
\affiliations{
    \textsuperscript{\rm 1}Bytedance Seed \quad \textsuperscript{\rm 2}Shanghao Jiao Tong University \quad \textsuperscript{\rm 3} Peking University\\
\vspace{8pt}

\{hyc, linyuan.0, wjc, zhangyuchen.zyc\}@bytedance.com, \\
zhanghanchong@sjtu.edu.cn,\quad 2200017806@stu.pku.edu.cn

}

\usepackage{bibentry}
\begin{document}

\maketitle

\begin{abstract}
    Existing large vision-language models (LVLMs) are largely limited to processing short, seconds-long videos and struggle with generating coherent descriptions for extended video spanning minutes or more. Long video description introduces new challenges, such as consistent character identification and plot-level descriptions incorporating both visual and audio information. To address these, we figure out audio-visual character identification, matching character names to each dialogue, as a key factor. We propose \systemnm, a system for generating dense descriptions of long videos, incorporating both low-level visual concepts and high-level plot information. \systemnm uses a multimodal large language model that integrates visual, audio, and text modalities to perform audio-visual character identification on minute-long video clips. The results are then fed into a LVLM to enhance consistency of video description. We validate our approach on movie description tasks and introduce \datasetnm, a dataset with dense descriptions for three-minute movie clips. To evaluate long video descriptions, we create \qanm, a large set of multiple-choice questions for \datasetnm test set. We assess descriptions by inputting them into GPT-4 to answer these questions, using accuracy as an automatic evaluation metric. Experiments show that \systemnm outperforms all open and closed-source baselines on \qanm, achieving 9.5\% higher accuracy than the strongest baseline, Gemini-1.5-pro, and demonstrating a +15.56\% advantage in human side-by-side evaluations. Additionally, incorporating audio-visual character identification from \systemnm improves the performance of all video description models, with Gemini-1.5-pro and GPT-4o showing relative improvement of 5.5\% and 13.0\%, respectively, in accuracy on \qanm. %Datasets and code are available at \link{https://github.com/hyc2026/StoryTeller}.
\end{abstract}
\begin{links}
    \link{Code}{https://github.com/hyc2026/StoryTeller}
\end{links}

\section{Introduction}

Generating detailed video descriptions is a core challenge in video understanding. While large vision-language models (LVLMs) have made significant progress~\cite{wang2024qwen2,lin2024vila,chen2024far,wang2024tarsier}, they remain limited to processing seconds-long videos and generating low-level visual concepts like objects, scenes and atomic actions. Real-world videos, such as movies, are much longer and require high-level information for effective description, such as plot development. Recent research works address long videos by segmenting them, generating descriptions for each segment, and combining them using large language models~\cite{zhang2024mm,han2023autoad,han2023autoad-ii}. However, these methods struggle to preserve character identity and overlook audio cues, making coherent, detailed, plot-level descriptions difficult to achieve.

Thus, accurate character identification and tracking across both visual and audio modalities are essential for high-quality long video descriptions. While existing studies~\cite{raajesh2024micap,ji2024ida} have improved the quality of video descriptions by tracking character identification via visual cues, a critical gap remains in linking dialogue with on-screen characters to identify speakers. Bridging this gap is essential for enhancing long video descriptions.

To address these issues, we propose a new task: given a video and its cast list with character photos and names, identify the speaker for each dialogue line in the video using both visual and auditory cues. If the character is in the cast, provide their name; otherwise, give a descriptive label. We refer to this task as \emph{audio-visual character identification}. This task is challenging as it requires integrating visual, auditory, and textual information to link dialogue lines to character identities. To tackle this, we developed a multimodal large language model equipped with both visual and auditory inputs. Due to the limitation of models with visual encoder in handling only short segments, it struggles to utilize global auditory information (e.g., some lines spoken by the same character through a long video). To address this, we introduce a global decoding algorithm that incorporates global information during inference to improve accuracy of audio-visual character identification.

Furthermore, we propose \systemnm, a novel system for generating dense descriptions of long videos, incorporating both basic visual concepts and high-level plot information with high consistency. The system consists of three components: a video segmentation module, an audio-visual character identification module, and an identity-aware description generation module, as illustrated in Figure~\ref{pipeline}.

A major challenge in long video description is the lack of training and test data. To address this, we introduce \datasetnm, a new dataset focused on movies—long videos with high information density. The videos in \datasetnm are from Movie101\cite{yue2023movie101} and Movie101v2~\cite{yue2024movie101v2}, comprising 187 movies split into 5,982 three-minute clips. Each clip is annotated with detailed storyline descriptions, subtitles, and character identity labels for each dialogue line. The storyline annotations ensure a complete understanding of the movie's plot. By training on this dataset, we aim to unleash the model's ability to generate plot-level descriptions. In addition, evaluating long video descriptions is also challenging. Besides human evaluation, we propose an automated evaluation method for \datasetnm's test set: \qanm. \qanm has 11 multiple-choice questions in average for each video, covering visual actions, character relationships, and plot details. We evaluate a long video description by inputting it into GPT-4~\cite{openai2023gpt4} to answer these questions, using the accuracy as an evaluation metric.

Based on \qanm and human side-by-side evaluation, we demonstrate that \systemnm outperforms Gemini-1.5-pro~\cite{reid2024gemini}, GPT-4o~\cite{gpt4o}, and several advanced open-source LVLMs~\cite{lin2024vila,li2024llava,wang2024qwen2,chen2023internvl} in generating descriptions for long videos. Specifically, the 7B model-based \systemnm achieves an accuracy in the \qanm evaluation that is 9.5\% higher than the best-performing baseline model, Gemini-1.5-pro. In human side-by-side evaluation, \systemnm shows an advantage of +15.56\% over Gemini-1.5-pro. Additionally, we verify the significant role of accurate audio-visual character identification on improving description. Incorporating audio-visual character identification results from \systemnm as input consistently improves the performance of various LVLMs, with Gemini-1.5-pro and GPT-4o
achieving relative accuracy gains of 5.5\% and 13.0\% on \qanm, respectively.

\section{Related Work}

\noindent{\bf Long Video Description.} Limited by the number of input frames, LVLMs are primarily effective for short video clips and typically generate descriptions focused on low-level visual concepts~\cite{wang2024qwen2,lin2024vila,chen2024far,wang2024tarsier,lin2023video,li2023videochat}. There remains a gap in describing longer videos, which may span several minutes or more, and in capturing higher-level content like character interactions and plot progression.Some research~\cite{zhang2024mm,han2023autoad,han2023autoad-ii} treats long videos as a series of short clips, incorporating previous descriptions when describing the current one, but this can lead to cumulative errors that affect consistency and coherence of the overall narrative~\cite{han2023autoad}. Another approach, Video ReCap~\cite{islam2024video}, offers a recursive model for varying video lengths. Nevertheless, for longer videos, this method tends to produce summaries, limiting its ability to generate detailed descriptions.

\noindent{\bf Audio-Visual Character Identification.} Character identification is essential for high-quality video descriptions~\cite{raajesh2024micap,ji2024ida}, particularly to maintain consistency across clips. While prior research has focused on visual-level identification~\cite{nagrani2018benedict,tapaswi2012knock}, leveraging audio cues to determine which character is speaking can straightforwardly improve descriptions. A closely related area of research is audio-visual diarization, which integrates both visual and auditory features to segment speakers in audio~\cite{xu2022ava,he2022end}. Nonetheless, current methods are still unable to identify the speakers' identities.

\begin{figure*}[h]
\centering
\vspace{-10pt}
\includegraphics[width=0.8\textwidth]{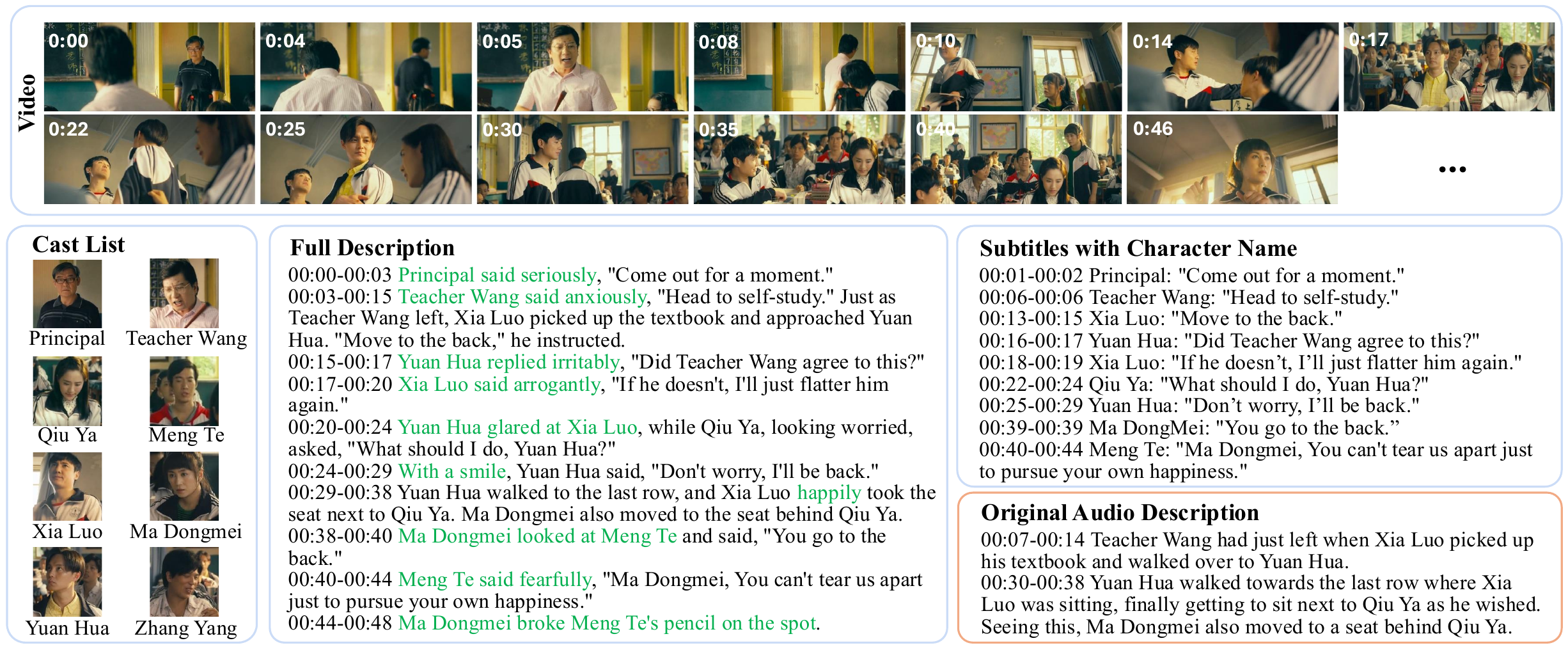}
\vspace{-5pt}
\caption{Example from \datasetnm: Each sample includes a 3-minute video, a detailed description, cast list, and subtitles with character names. We compare the full description with audio description (from Movie101); green text highlights additional details in full description, such as character identification, emotions, and activities.}
\label{dataset}
%\vspace{-5pt}
\end{figure*}

\section{\datasetnm Dataset}

\begin{table*}[htbp]
%\vspace{-12pt}
  \centering
  \resizebox{0.8\linewidth}{!}{%
  \begin{tabular}{lccccccc}
    \toprule
        \multicolumn{1}{c}{Dataset} & Video Source & Description Style & \# Movie & \# Clip & $L_v$ & $L_t$ & $N_{\text{char}}$\\
    \midrule
     MovieNet~\cite{huang2020movienet}  & movie & - & 1100 & 42K & 428.0 & - & - \\ 
    % \midrule
     ActivityNet Captions~\cite{krishna2017dense}  & ActivityNet & activity description & - & 20k & 180.0 & 49.2 & - \\
     YouCook2~\cite{ZhXuCoAAAI18}  & Youtube cook video & activity description & - & 2k & 316.2 & 67.8 & - \\
     ViTT~\cite{huang2020multimodal}  & Youtube video & activity tag & - & 8k & 282.0 & 21.1 & - \\
     CMD-AD~\cite{han2024autoad}  & movie & audio description & 1432 & 101k & 2.8 & 9.1 & - \\
     Movie101~\cite{yue2023movie101} & movie & audio description & 101  & 14.1k & 20.4 & \underline{80.7} & 2.0 \\
     Movie101-v2~\cite{yue2024movie101v2} & movie & audio description & 203 & 46k & 12.8 & 39.1 & 1.9 \\
     \midrule
     \datasetnm & movie & full description & 187 & 6k & 180.0 & 565.0 & 4.3 \\
     \bottomrule
  \end{tabular}}
  \vspace{-5pt}
  \caption{Data statistics for \datasetnm compared with other video description datasets. $L_{v/t}$: average clip duration (seconds) or text length (words;  Chinese characters underlined); $N_{\text{char}}$: average number of characters per clip.}
  \label{tab:statistics}
  \vspace{-10pt}
\end{table*}

\begin{table}[htbp]
%\vspace{-12pt}
  \centering
  \resizebox{1.0\linewidth}{!}{%
  \begin{tabular}{lccccc}
    \toprule
        \multicolumn{1}{c}{Dataset} & \begin{tabular}[c]{@{}c@{}}full \\description\end{tabular} & \begin{tabular}[c]{@{}c@{}}character \\photo\end{tabular} &  subtitle & \begin{tabular}[c]{@{}c@{}}speaker \\name\end{tabular} & \begin{tabular}[c]{@{}c@{}}audio \\description\end{tabular} \\
    \midrule
     MovieNet~\cite{huang2020movienet}  & \ding{55} & \ding{51} & \ding{51} & \ding{55} & \ding{55} \\ 
    % \midrule
     MovieQA~\cite{tapaswi2016movieqa} & \ding{55} & \ding{55} & \ding{51} & \ding{55} & \ding{51} \\
     CMD-AD~\cite{han2024autoad} & \ding{55} & \ding{51} & \ding{55} & \ding{55} & \ding{51} \\
     Movie101~\cite{yue2023movie101} & \ding{55} & \ding{51} & \ding{51} & \ding{55} & \ding{51} \\
     Movie101-v2~\cite{yue2024movie101v2} & \ding{55} & \ding{51} & \ding{51} & \ding{55} & \ding{51} \\
     \midrule
     \datasetnm & \ding{51} & \ding{51} & \ding{51} & \ding{51} & \ding{55} \\
     \bottomrule
  \end{tabular}}
  \vspace{-5pt}
  \caption{Annotation comparison between \datasetnm and related movie datasets. Note that the subtitles provided by Movie101 and Movie101v2 are generated using ASR and \datasetnm manually revised them.}
  \label{tab:annotation}
  \vspace{-15pt}
\end{table}

\datasetnm is a long video description dataset based on movies from Movie101~\cite{yue2023movie101} and Movie101v2~\cite{yue2024movie101v2}. We choose movies because they contain rich and diverse information. A high-quality movie description should capture not only fine-grained details such as scenes, characters, and actions but also higher-level aspects like character relationships, plot development, and causal links between events (see Figure~\ref{dataset}). 

The \datasetnm task is related to audio description for movies~\cite{han2023autoad, han2023autoad-ii, han2024autoad}, which narrates non-dialogue segments, and dense video captioning~\cite{krishna2017dense,chen2020continuous}, which generates temporally localized descriptions. However, \datasetnm goes further by offering comprehensive, plot-level description, integrating both visual and audio information. We refer to this as \emph{full description}. Figure~\ref{dataset} illustrates the difference between a full description and the original audio description from Movie101. In Table~\ref{tab:statistics}, we report the statistics of \datasetnm and compare them with those of the other video description datasets, emphasizing its unique features: detailed descriptions, long video durations, and a large number of characters. Additionally, we provide a comparison of the annotations across different movie datasets in Table~\ref{tab:annotation}.

\begin{table*}[t]
    \centering
    \scalebox{0.8}{
        \begin{tabular}{m{1.5cm}m{5.5cm}m{5.5cm}m{5.5cm}}
\toprule[1pt]
            \multicolumn{1}{l}{\textbf{Type}} & \multicolumn{1}{l}{\textbf{Question}} & \multicolumn{2}{l}{\textbf{Options}} \\
\midrule[0.5pt]

            Action & What does Jiang Feng do immediately after he tells Hui Lan that he saw `her'? & A. Leaves the hut\newline C. Calls someone on the phone & B. Cries\newline
D. Prepares for surgery\\
\midrule[0.5pt]
%Action & What does Jiang Feng do before going to bed at night?&
%A. Drinks water\newline
%C. Takes a sleeping pill & B. Writes in a diary\newline
%D. Takes a shower\\
%\midrule[0.5pt]
%            Character & Who suggests Jiang Feng to rethink about his decision to undergo the memory erasing surgery? & A. Zhang Daichen\newline
%            C. Hui Lan\newline &  B. The boss\newline
%            D. The person on the other side of the phone\\
%\midrule[0.5pt]
Character & How does Jiang Feng react when Hui Lan decides to leave? &
A. He insists on accompanying her\newline
C. He gets angry at her decision & B. He advises her not to do so\newline
D. He feels relieved\\
\midrule[0.5pt]
Plot & What must happen within 72 hours of the first memory reload for Jiang Feng? &
A. Another memory extraction surgery\newline
C. Restore all his memories & B. Take a special pill\newline
D. Make a reservation\\
%\midrule[0.5pt]
%Plot & Why does Hui Lan unplug the phone line? & 
%A. To prevent Jiang Feng from making a call\newline
%C. As per Jiang Feng's suggestion & B. Because she is scared\newline\newline
%D. To charge her phone\\
\bottomrule[1pt]
        \end{tabular}
    }
    \vspace{-5pt}
    \caption{Examples of Action, Character, and Plot Questions in \qanm.}
    \label{movie_qa_examples}
\vspace{-10pt}
\end{table*}

\subsection{Movie Source}

We construct \datasetnm based on Movie101 and Movie101v2, which together comprise 203 movies. Since both sources contain audio descriptions integrated into the audio tracks, we collect the original editions to obtain the original audio tracks. After review, we select 187 movies for our dataset, excluding 16 movies because their original editions differed significantly from their barrier-free versions. Each selected movie is segmented into 3-minute clips. The final \datasetnm dataset includes 5,210 / 140 / 632 clips in
the training / development / testing split, totaling 300 hours. Each split contains video clips from different movies.

\begin{figure}[h]
\centering
\includegraphics[width=0.48\textwidth]{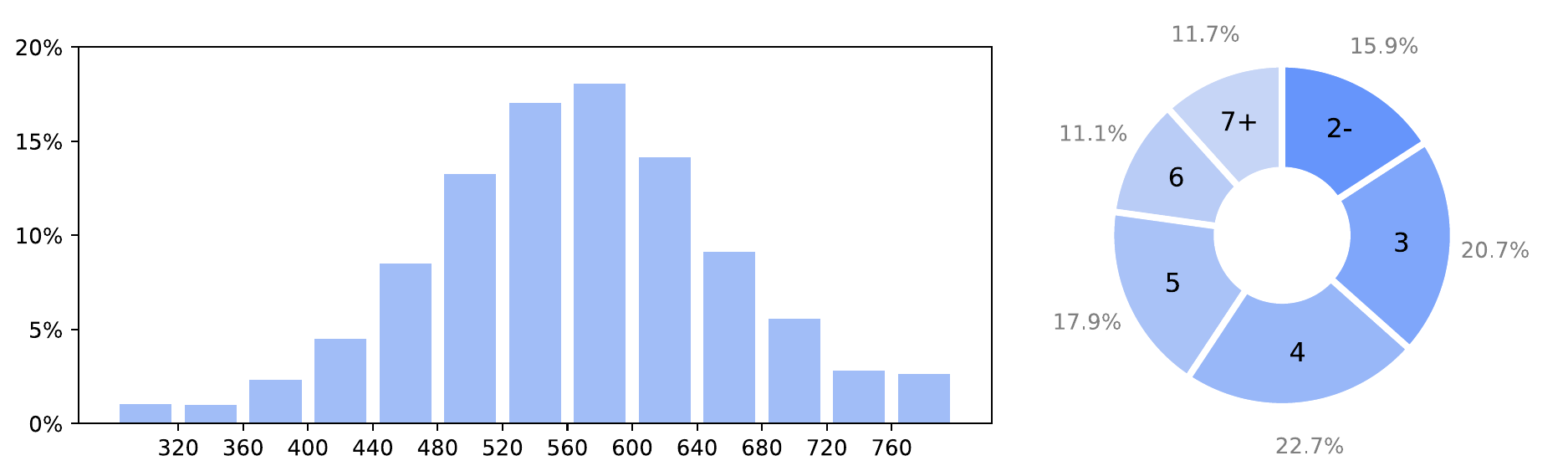}
\vspace{-15pt}
\caption{Distribution of description length (left) and character count (right) in \datasetnm.}
\label{statistic1}
\vspace{-10pt}
\end{figure}
% Besides video and audio, we also collect subtitle for each movie.

\subsection{Annotation Process}

The annotation process includes movie descriptions, character photos, and audio-visual character identification. 

\noindent{\bf Movie description.} We annotate each movie clip with detailed description. To improve efficiency, annotators use audio descriptions and ASR data from Movie101 and Movie101v2 as references. Annotators are tasked with providing the descriptions of each event in the video, along with the corresponding start and end timestamps. The descriptions should include details about scenes, characters, actions, dialogues, and plot development. In addition, annotators revise ASR data to generate subtitles. %The aim is to create a coherent long-form description of the video. 

\noindent{\bf Character photos.} Annotators identify main characters and select a face for each in each movie clip, creating a cast list with names and corresponding faces.

\noindent{\bf Audio-visual character identification.} Annotators identify the speaker for each dialogue line in the clips. If the speaker is in the cast list, their character name is assigned; otherwise, a descriptive name is used.

See annotation guidelines and annotator information in Appendix~\ref{annotation_guidelines}. Figure~\ref{dataset} provides an example in \datasetnm dataset. Figure~\ref{statistic1} shows the statistics for descriptions and characters in \datasetnm.

\begin{figure}[h]
\centering
\vspace{-5pt}
\includegraphics[width=0.48\textwidth]{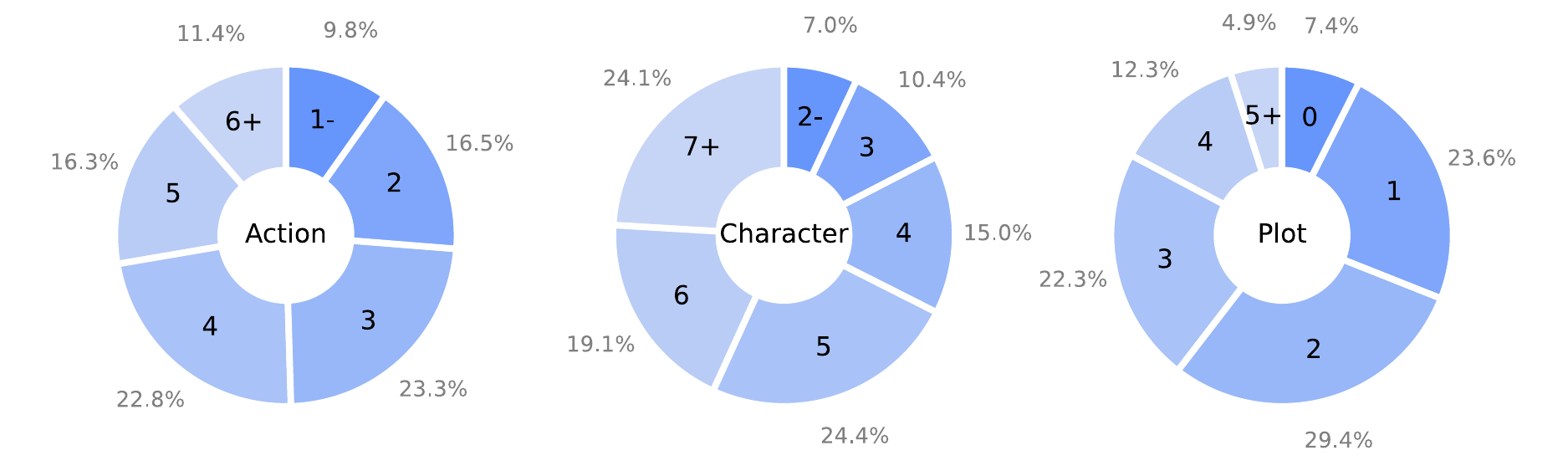}
\vspace{-20pt}
\caption{The distribution of the three types of QA counts across each clip in the \datasetnm test set.}
\label{statistic2}
\vspace{-10pt}
\end{figure}

\subsection{Evaluation of Description}

Evaluating long video descriptions is challenging due to the numerous details they contain, which increases evaluation costs. To address this, we develop an automatic evaluation method called \qanm for the \datasetnm test set. On average, each 3-minute video in \qanm is accompanied by 38 multiple-choice questions focusing on visual actions, character relationships, and plot development. During evaluation, we employ GPT-4~\cite{openai2023gpt4} to answer these questions given the long description as context, and the QA accuracy serves as an automated metric for assessing the quality of the description. The prompt template is shown in Appendix~\ref{templates}. Table~\ref{movie_qa_examples} shows examples in \qanm.

\begin{figure*}[h]
\centering
\includegraphics[width=0.9\textwidth]{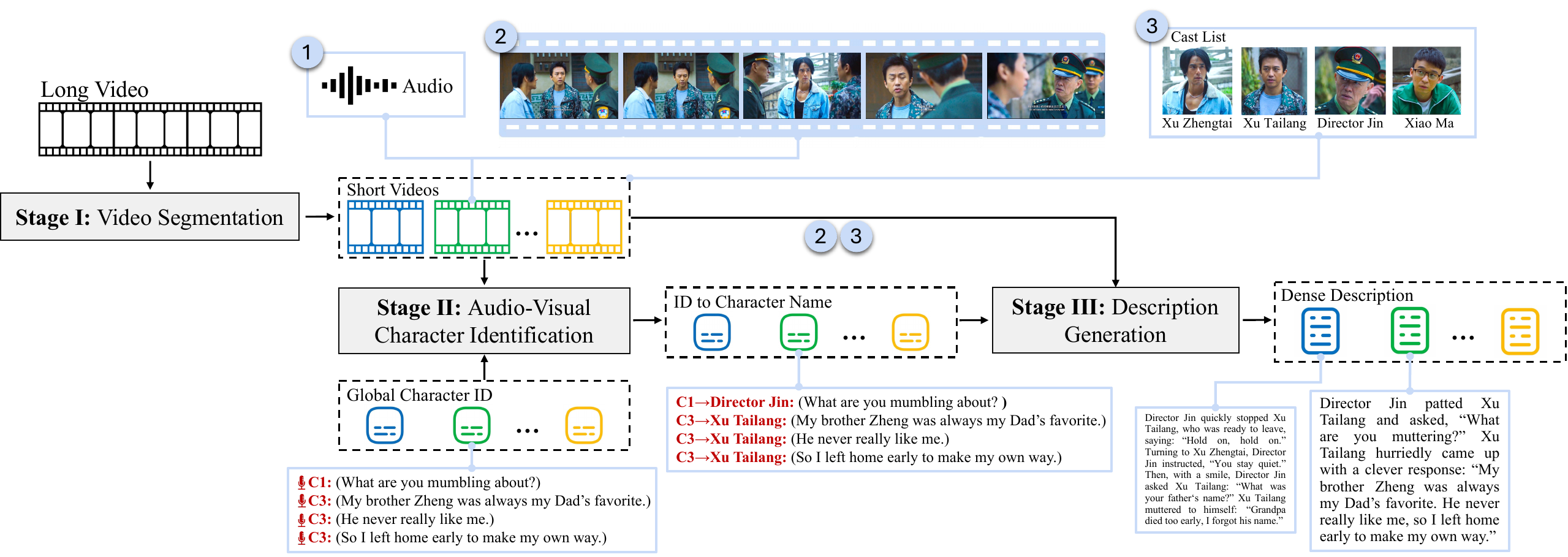}
\caption{
Pipeline of \systemnm, consisting of three modules: Stage I - Video Segmentation, where videos are divided into short clips that are relatively independent and internally complete; Stage II - Audio-Visual Character Identification, where characters are identified for each dialogue line using both audio and visual cues; and Stage III - Description Generation, where detailed descriptions are generated for each short clip, ultimately producing a consistent narrative for the entire long video.}
\label{pipeline}
\vspace{-10pt}
\end{figure*}

\noindent{\bf Creation for \qanm} We identify three key types of information for evaluating long video descriptions: 1) {\bf Action}: actions, their sequence, and causal relationships; 2) {\bf Character}: roles, relationships, and their evolution; 3) {\bf Plot}: causes, character motivations, and event connections. We generate QA pairs for each type, as detailed in Appendix~\ref{movie_qa_pipeline}. Figure~\ref{statistic2} shows the distribution of QA counts across each clip in the \datasetnm test set.

\noindent{\bf Validation for \qanm} To evaluate the quality of \qanm, we sample 100 QA pairs, checking if each question is answerable based on the corresponding video and if the correct answer is among the options. The qualification rate is 98\%. Then, we evaluate the GPT-4 evaluator by sampling 100 (description, question) pairs, with descriptions from both baselines in the Experiments section and \systemnm. The authors answer each question, and we measure GPT-4's agreement with the authors, which is 95\%.

\begin{figure*}[h]
\centering
\vspace{-10pt}
\includegraphics[width=0.8\textwidth]{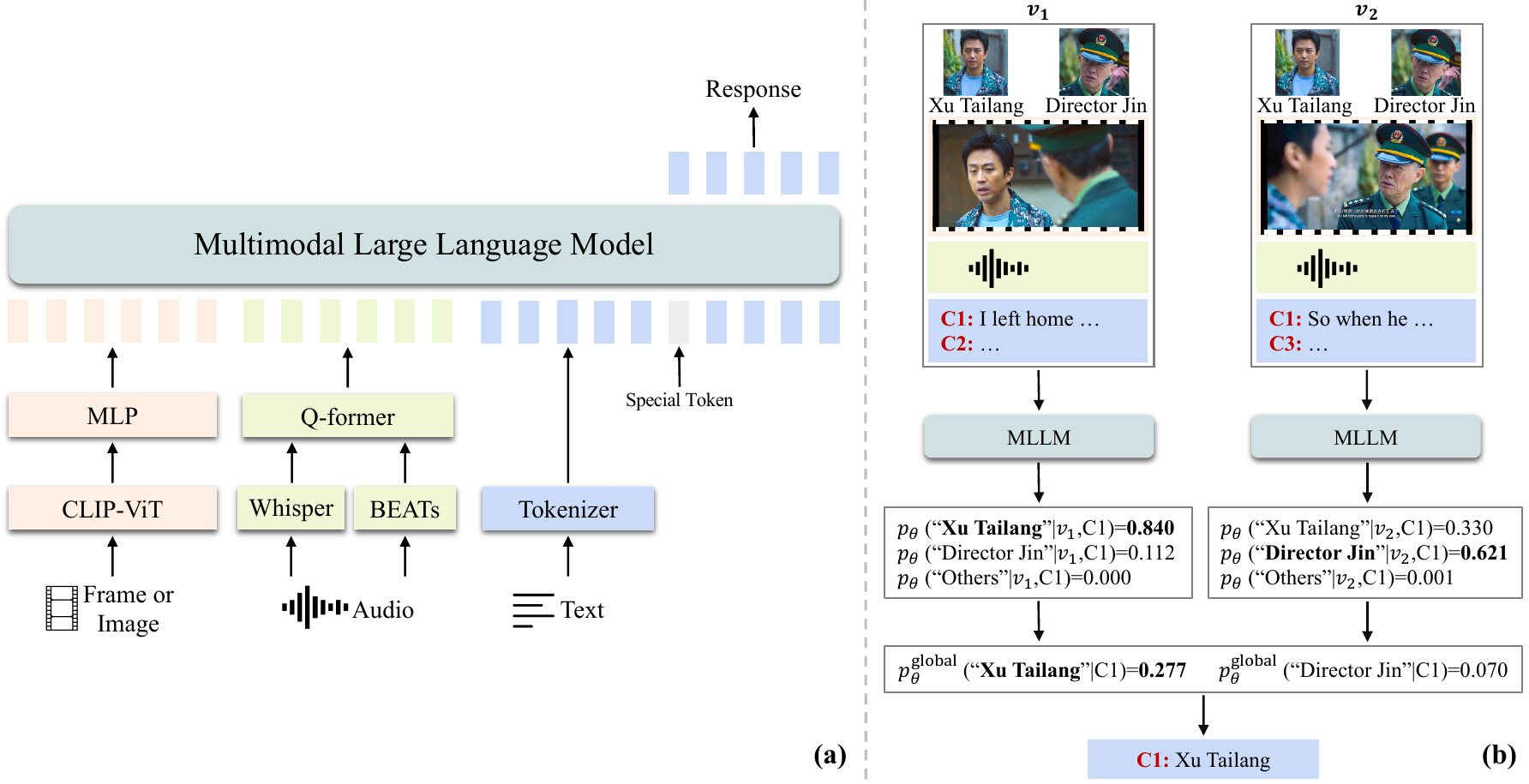}
\caption{Audio-visual character identification module. (a) A multimodal large language model integrating visual, audio and text inputs is employed to identify characters in seconds-long videos. (b) Global Decoding: During inference, consistent mapping of global IDs across different short clips ensures the same global ID is assigned the same character name.}
\label{decode}
\vspace{-10pt}
\end{figure*}

\section{\systemnm Long Video Description System}

\subsection{Overview}

\systemnm is a fully automated system for generating descriptions of long videos. As shown in Figure~\ref{pipeline}, it consists of three main modules: 1) the \emph{video segmentation} module divides the long video into multiple short clips, each lasting several seconds, while preserving the integrity and relative independence of these clips, as detailed in Appendix~\ref{segmentation_method}; 2) the \emph{audio-visual character identification} module, which employs a dual mechanism, local and global, to accurately identify characters throughout the long video. Locally, a multimodal large language model (MLLM) integrates audio, visual, and text modalities to achieve character identification using both audio and visual cues within each short clip. Globally, during inference, a global decoding algorithm processes multiple short clips associated with the same character simultaneously, combining their local results from the MLLM to enhance the overall identification accuracy; 3) the \emph{description generation} module utilizes an LVLM to produce a detailed description. By incorporating the audio-visual character identification as input, the LVLM generates coherent video descriptions that span the entire long video, resulting in a plot-level dense narrative.

\subsection{Audio-Visual Character Identification}

The audio-visual character identification module determines speakers identity in video clips by integrating audio, visual and textual information. In particular, given a video clip (frames, audio and subtitles) and a cast list (character photos and names), the module assigns each dialogue line to its corresponding character. If the speaker is in the cast list, their name is provided; otherwise, a descriptive identity (e.g., `a police officer' or `a little girl') is used. This task presents several challenges: 1) relying solely on visual information is often insufficient, especially when multiple characters are on screen or the speaker is not visible; 2) audio cues in seconds-long clips are typically isolated, making them less effective for character identification. However, in longer segments, audio can be leveraged to extract speaker features across different clips, such as identifying if the same person speaks in multiple segments. Models that are limited to short segments often struggle to incorporate such global features effectively.

To address the aforementioned challenges, we design a novel audio-visual character identification mechanism incorporating a MLLM equipped with both audio and video modalities. We also develop a global decoding algorithm to leverage global audio information for more accurate character identification. Specifically, as illustrated in Figure~\ref{pipeline}, we begin by performing audio diarization on the entire video, segmenting the dialogue, and assigning IDs to different speakers (e.g., C1, C2, etc.), as detailed in Appendix~\ref{global_id_genetation}. The subtitles with global IDs are then fed into the MLLM to identify the corresponding character names for each global ID. The global decoding mechanism processes short video clips in parallel, ensuring that global IDs in different clips are consistently mapped to the same name, thereby improving identification accuracy. Details are provided below. %Details on the implementation of the MLLM and the global decoding algorithm are provided below.

\noindent{\bf Model Architecture and Training.} We integrate a MLLM capable of processing both visual and auditory data, as depicted in Figure~\ref{decode}. For visual inputs, each image or video frame is encoded by CLIP-ViT~\cite{radford2021learning}, followed by mapping into the token embedding space of the downstream LLM through a multi-layer perceptron (MLP). These tokens are then fed into the LLM. The audio inputs are processed by dual auditory encoders~\cite{tang2023salmonn}, a speech encoder from Whisper-Large-v2~\cite{radford2023robust} and a non-speech BEATs audio encoder~\cite{chen2022beats}. The outputs from the two audio encoders are concatenated at the frame level. Subsequently, a window-level Q-former~\cite{li2023blip} align the combined audio representation to the LLM's input space.

We initialize the visual module and language model from the Tarsier-7b~\cite{wang2024tarsier}, a large vision-language model demonstrating SOTA performance on various video understanding benchmarks. Our training protocol consists of three phases: (1) pre-training the audio module, (2) fine-tuning on tasks including audio diarization, character identification, recognition and tracking, and (3) fine-tuning for audio-visual character identification using \datasetnm training set. Details on the dataset used in each phase and the complete training methodology are provided in Appendix~\ref{training_detail}.

\noindent{\bf Global Decoding.} The MLLM assigns names to each global character ID. First, these names come from a set $\mathcal{N}$, including all names from the cast list as well as a special item ``Others", representing characters not in the cast list. When the model outputs ``Others", it will further provide a descriptive name. Second, the model outputs a JSON dictionary, where each key is a global character ID within the input short clip, and each corresponding value is the assigned name. Keys are listed in ascending order of global character IDs.

A given global character ID may appear simultaneously in multiple short video clips. If we independently assign names to the ID across different clips, conflicts may arise where the same character ID is assigned different names in different clips. To address this, we decode the results from all clips containing the same global character ID in parallel.

Specifically, we consider a multimodal LLM parameterized by $\theta$, which assigns names to each global character ID in a fixed ascending order (i.e., C1, C2, ...). For a given global character ID $x$ appearing in $m$ short video clips, denoted as $v_1, \cdots, v_m$, the probability that the model assigns a name $y\in\mathcal{N}$ to $x$ in clip $v_k$ is given by $p_\theta(y|v_k, x)$, which can be computed as:
\begin{equation}
    p_\theta(y|v_k, x)=\prod_{t}p_\theta(y_{t}|v_k, x, y_{<t}),
\end{equation}
where $y_{t}$ is the $t$-th token in the response and $y_{<t}$ is tokens in the response before $y_{t}$.

\begin{table*}[htbp]
\vspace{-10pt}
\centering
\scalebox{0.8}{
\begin{tabular}{lccccccc}
\toprule[1pt]
\multicolumn{1}{c}{\multirow{2}{*}{\textbf{Model}}} & \multicolumn{4}{c}{\textbf{\qanm Accuracy~$\uparrow$}} & \multirow{2}{*}{\textbf{CIDEr~$\uparrow$}} & \multirow{2}{*}{\textbf{R@1/5~$\uparrow$}} & \multirow{2}{*}{\textbf{CRITIC~$\uparrow$}} \\
& \textbf{Character} & \textbf{Action} & \textbf{Plot} & \textbf{Total}\\
\midrule[0.5pt]
Gemini-1.5-pro & 0.578 & 0.501 & 0.534 & 0.544 & 0.035 & 0.231 & 0.497 \\
GPT-4o & 0.517 & 0.479 & 0.528 & 0.507 & 0.000 & 0.318 & 0.370 \\
% Tarsier-7B & 0.599 & 0.509 & 0.588 & 0.567 \\
VILA1.5-8B & 0.561 & 0.459 & 0.540 & 0.524 & 0.120 & 0.339 & 0.608 \\
LLaVA-OneVision-7B & 0.557 & 0.454 & 0.540 & 0.520 & 0.199 & 0.339 & 0.615 \\
Qwen2-VL-7B & 0.549 & 0.468 & 0.549 & 0.523 & 0.084 & 0.340 & 0.594 \\
InternVL2-8B & 0.535 & 0.448 & 0.506 & 0.501 & 0.168 & 0.336 & 0.573 \\
\midrule[0.5pt]
\textbf{\systemnm} & \textbf{0.676} & \textbf{0.583} & \textbf{0.644} & \textbf{0.639} & \textbf{0.237} & \textbf{0.342} & \textbf{0.630} \\
\bottomrule[1pt]
\end{tabular}
}
\caption{Results of video descriptions generated by \systemnm and baselines, evaluated using \qanm and traditional metrics.}
\vspace{-10pt}
\label{overall_result}
\end{table*}

Then, the global probability of $x$ corresponding to name $y$ is defined as
\begin{equation}
    p^{\rm global}_\theta(y|x)=\prod_{k=1}^m p_\theta(y|v_k,x).
\end{equation}
The name assigned to the global character ID $x$ is then determined as $\arg\max_{y\in\mathcal{N}} p^{\rm global}_\theta(y|x)$.

For example, as shown in Figure~\ref{decode}, both videos $v_1$ and $v_2$ contain lines from character $C1$. In $v_2$, although Xu Tailang speaks, the camera focuses entirely on silent Director Jin, leading the model assigns a higher probability to Director Jin being the speaker. However, the model still assigns over 30\% confidence to Xu Tailang. By integrating information from the entire video globally, $C1$ is ultimately mapped to Xu Tailang, correcting the erroneous prediction in $v_2$.

In addition, when calculating the probability of the model generating the special item ``Others", only the token probability for ``Others" is considered, excluding any descriptive name tokens. This approach is based on our observation that the model consistently achieves high accuracy in generating descriptive names, but during inference, it may produce different yet valid descriptions for the same individual, such as `A policeman' and `A man in police uniform.' Although these descriptions vary, they do not impact the final video descriptions, as the language model can correctly identify that they refer to the same person.

\subsection{Description Generation}

After identifying the character, we proceed to generate the final description. At this stage, the video, cast list, and subtitles with character names are provided to the model. Audio input is excluded to ensure compatibility with most existing open-source or closed-source large vision-language models. For open-source models, we use video description data from the \datasetnm training set for supervised fine-tuning. During this stage, both visual adapter and LLM are trained.

\begin{figure*}[h]
\centering
\vspace{-10pt}
\includegraphics[width=0.95\textwidth]{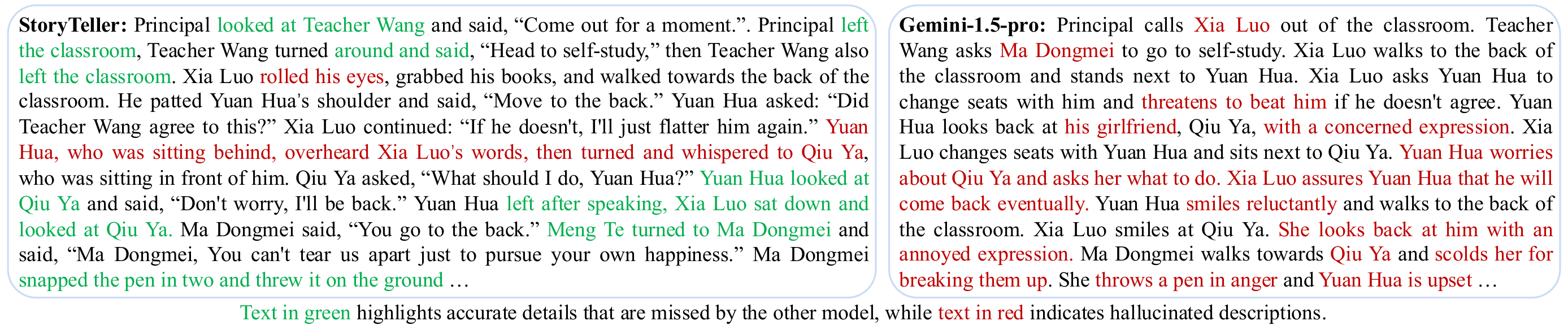}
\vspace{-10pt}
\caption{Comparison of StoryTeller and Gemini-1.5-pro on a video example featuring multiple characters (same as Figure~\ref{dataset}).}
\label{casestudy}
\end{figure*}

\section{Experiments}

\subsection{Baselines}
We compare our method against two types of baselines:

\begin{itemize}
    \item {\bf Closed-Source LVLMs:} We compare our method with Gemini-1.5-pro~\cite{reid2024gemini} and GPT-4o~\cite{gpt4o}. For both models, we perform prompt engineering to generate detailed video descriptions that incorporate both fine-grained details and high-level plot information. The prompt templates are provided in Appendix~\ref{templates}.
    \item {\bf Open-Source LVLMs:} We also compare our method with open-source LVLMs that demonstrate strong video understanding capabilities, including VILA1.5-8B \cite{lin2024vila}, LLaVA-OneVision-7B \cite{li2024llava}, Qwen2-VL-7B \cite{chen2024qwen2}, and InternVL2-8B \cite{chen2023internvl, chen2024far}, all fine-tuned on the \datasetnm training set for fairness. 
\end{itemize}
For all baselines, we provide the video, cast list and subtitles as input. Specifically, for Gemini-1.5-pro, we input the entire 3-minute video, while for other models, we divide the video into 10-second clips and sample 8 frames from each. Finally, we concatenate the descriptions from each clip to form a complete description of the entire 3-minute video.

\subsection{Experimental Results}

First, we evaluate the descriptions generated by our method and the baselines on the \qanm dataset. As shown in Table~\ref{overall_result}, \systemnm outperforms all baselines on \qanm. Notably, \systemnm achieves a 9.5\% higher accuracy than the strongest baseline, Gemini-1.5-pro. Table~\ref{casestudy} shows a video example featuring multiple characters, where \systemnm effectively tracks their relationships, resulting in description with fewer hallucination compared to Gemini-15-pro. When analyzing performance across different question types in \qanm, \systemnm outperforms all baselines on action, character and plot questions, indicating that \systemnm provides more accurate descriptions of both low-level visual details and high-level plot information. 

\begin{table}[ht]
\vspace{-5pt}
\centering
\scalebox{0.8}{
\begin{tabular}{ccccc}
\toprule[1pt]
\textbf{Baseline} & \textbf{Win} & \textbf{Same} & \textbf{Loss} & \textbf{Advantage}\\
\midrule[0.5pt]
% Align & 0.717~(0.662) & 0.745~(0.752) \\
% Align~\&~Pretrain & 0.759~(0.700) & 0.791~(0.811) \\
Gemini-1.5-pro & 40.89\% & 33.78\% & 25.33\% & +15.56\% \\
VILA1.5-8B & 56.11\%  & 30.03\% & 13.86\% & +42.25\%\\
\bottomrule[1pt]
\end{tabular}
}
\vspace{-5pt}
\caption{Human side-by-side evaluation results on the \datasetnm test set (StoryTeller vs. baseline). \textbf{Win}, \textbf{Same}, and \textbf{Loss} refer to StoryTeller outcomes. \textbf{Advantage} is calculated as \textbf{Win} – \textbf{Loss}.}
\label{human_evaluation}
\vspace{-10pt}
\end{table}

\begin{table*}[htbp]
\centering
\scalebox{0.8}{
\begin{tabular}{lcp{1.6cm}<{\centering} p{1.3cm}<{\centering} p{1.3cm}<{\centering} p{1.1cm}<{\centering} p{1.6cm}<{\centering} p{1.3cm}<{\centering}}
\toprule[1pt]
\multicolumn{1}{c}{\multirow{2.6}{*}{\textbf{Model}}} & \textbf{Baseline} & \multicolumn{2}{c}{\textbf{+ Segmentation Method}} & \multicolumn{2}{c}{\textbf{+ Characher Names}} & \multicolumn{2}{c}{\textbf{+ Descriptive Names}} \\
\cmidrule[0.5pt](lr){2-2} \cmidrule[0.5pt](lr){3-4} \cmidrule[0.5pt](lr){5-6} \cmidrule[0.5pt](lr){7-8}
& \textbf{Acc} & \textbf{Acc} & \textbf{Impv} & \textbf{Acc} & \textbf{Impv}  & \textbf{Acc} & \textbf{Impv} \\
\midrule[0.5pt]
Gemini-1.5-pro\cite{reid2024gemini} & - & 0.544 & - & 0.566 & 4.0\% & 0.574 & 1.4\% \\
GPT-4o\cite{gpt4o} & 0.507 & 0.519 & 2.4\% & 0.564 & 8.7\% & 0.573 & 1.6\% \\
VILA1.5-8B\cite{lin2024vila} & 0.524 & 0.566 & 8.0\% & 0.6 & 6.0\% & 0.605 & 0.8\% \\
LLaVA-OneVision-7B\cite{li2024llava} & 0.52 & 0.561 & 7.9\% & 0.593 & 5.7\% & 0.598 & 0.8\% \\
% Qwen2-VL-7B\cite{chen2024qwen2} & & & & & & & \\
Qwen2-VL-7B\cite{chen2024qwen2} & 0.523 & 0.555 & 6.1\% & 0.597 & 7.6\% & 0.601 & 0.7\% \\
InternVL2-8B\cite{chen2023internvl,chen2024far} & 0.501 & 0.541 & 8.0\% & 0.578 & 6.8\% & 0.582 & 0.7\% \\
\textbf{\systemnm} & 0.567 & 0.597 & 5.3\% & 0.631 & 5.7\% & 0.639 & 1.3\% \\
\midrule[0.5pt]
Average & 0.524 & 0.555 & 6.3\% & 0.590 & 6.4\% & 0.596 & 1.0\% \\
\bottomrule[1pt]
\end{tabular}
}
\vspace{-5pt}
\caption{Evaluation of \systemnm as a general framework for improving video descriptions across vairous LVLMs. The performance on \qanm when key components are incrementally added: video segmentation module, integration of the audio-visual character identification (with `Others' label for unlisted characters), and the use of descriptive names. The baseline approach involves segmenting videos into 10-second clips and generating descriptions without audio-visual character identification.}
\label{ablation}
\vspace{-10pt}
\end{table*}

We continue to conduct a human side-by-side evaluation, considered the gold standard for assessing video descriptions. We compare \systemnm against Gemini-1.5-pro and VILA1.5-8B, separately representing the best closed-source and open-source baselines. In this evaluation, we randomly selected 100 3-minute videos from the \datasetnm test set and asked annotators to compare the descriptions generated by two models, collecting their preferences. Due to the difficulty of directly comparing 3-minute videos, annotators viewed them in 20-second segments and provided preferences for the corresponding descriptions. We calculated the win rate at the 20-second segment level. The results, presented in Table~\ref{human_evaluation}, align with our automatic evaluation findings. Descriptions generated by \systemnm significantly outperforms those generated by Gemini-1.5-pro and VILA1.5-8B. Specifically, \systemnm demonstrate a +15.56\% advantage over Gemini-1.5-pro and a +42.25\% advantage over VILA1.5-8B.

Besides StoryQA, we report results from traditional evaluation metrics, including CIDEr~\cite{vedantam2015cider}, R@1/5~\cite{han2023autoad-ii} and CRITIC~\cite{han2024autoad}, as shown in Figure~\ref{overall_result}. We should note that traditional metrics (e.g., BLEU, ROUGE, CIDEr) rely on n-gram overlap, often penalizing accurate descriptions with different wording. Embedding-based metrics like R@1/5 improve semantic evaluation but still struggle with long, detailed content. This limitation is evident in Table~\ref{overall_result}, where Gemini-1.5-pro scores lower than VILA1.5-8B despite outperforming it in human side-by-side evaluations. At last, as modern video description models are largely LLM-based and generally produce fluent outputs, we prioritize evaluating content accuracy. StoryQA serves as an effective and scalable automated evaluation approach.

The system design of \systemnm is a general framework applicable to various LVLMs to enhance their long video descriptions. Table~\ref{ablation} presents the additive contributions of each component for various LVLMs. First, replacing the direct division into 10-second clips with our video segmentation module consistently improves description accuracy across all models, resulting in an average relative increase of 6.3\% in accuracy on the \qanm. Second, utilizing the audio-visual identification results generated by our model as input for the LVLMs, we observe an average relative accuracy improvement of 6.4\%. This demonstrates that our model's audio-visual identification results is effective for enhancing video descriptions across various LVLMs. Lastly, we observe that replacing `Others' with descriptive names generated by our model further improves video descriptions, consistently benefiting all models. Notably, under our framework, Gemini-1.5-pro and GPT-4o achieve relative improvements of 5.5\% and 13.0\% in accuracy, respectively.

\begin{table}[htbp]
\centering
\scalebox{0.9}{
\begin{tabular}{lc}
\toprule[1pt]
\textbf{Method} & \textbf{Accuracy} \\
\midrule[0.5pt]
% Align & 0.717~(0.662) & 0.745~(0.752) \\
% Align~\&~Pretrain & 0.759~(0.700) & 0.791~(0.811) \\
w/o subtitle & 0.741 \\
w/o audio & 0.722 \\
w/o video & 0.430 \\
\midrule
w/o sft on audio and visual tasks & 0.745 \\
w/o global decoding & 0.759 \\
\midrule[0.5pt]
\textbf{Ours} & \textbf{0.791}\\
\bottomrule[1pt]
\end{tabular}
}
\vspace{-5pt}
\caption{Ablation study results for audio-visual character identification accuracy on \datasetnm test set.}
\label{character_identification_ablation}
\vspace{-15pt}
\end{table}

At last, we conduct ablation study to evaluate the necessity of multi-modal inputs, fine-tuning for audio and visual tasks (audio diarization, character recognition, identification, and tracking), as well as effect of global decoding within the audio-visual identification module. We assess the accuracy of audio-visual character identification on the \datasetnm test set, as shown in Table~\ref{character_identification_ablation}. It  indicates that combining multi-modal inputs is crucial for character identification. Specifically, removing subtitle, audio or video inputs results in a reduction in accuracy by 5.0\%, 6.9\% and 36.1\%, respectively. In addition, fine-tuning the audio and visual tasks improves performance by 4.6\%, while global decoding contributes 3.2\% improvement. 

\section{Conclusion}

In this paper, we introduce \systemnm, a system for producing detailed, plot-level descriptions, composed of three modules: video segmentation, audio-visual character identification, and description generation. We introduce a new dataset, \datasetnm, with 5,982 three-minute movie clips from 187 movies. To effectively and efficiently evaluate long video descriptions, we present \qanm, a large-scale multiple-choice question-answering dataset aligned with the \datasetnm test set. We assess descriptions by inputting them into GPT-4 to answer these questions, using accuracy as an automatic evaluation metric. Experiments demonstrate that \systemnm outperforms all baselines, surpassing  Gemini-1.5-pro by 9.5\% in accuracy on \qanm and achieving a +15.56\% advantage in human side-by-side evaluations. 

\bibliography{aaai2026}

\appendix

\section{Annotation Details}\label{annotation_guidelines}

\subsection{Annotation Guidelines for Movie Description}

The task involves providing a detailed annotation of a 3-minute movie clip, including start and end times for each event. Describe the main events, scenes, characters, actions, dialogues, and plot development to create a cohesive narrative. Ensure the descriptions are logically coherent and enable full understanding of the video through text alone. Please follow the provided guidelines.

\begin{itemize}
\item When generating detailed descriptions, include the corresponding start and end timestamps for each event.
\item When describing key dialogues in the movie clip, provide the content of the dialogue, the manner of speaking, and any associated activities.
\item We provide the audio description and ASR of the movie clip. However, ensure that you label any missing key events, activities, or audio details not captured by these resources.
\item Provide fine-grained descriptions by segmenting longer descriptions that encompass multiple events. Each shorter description should have its own corresponding timestamps.
\item Label the subtitles of the movie clip, including their start and end times.
\end{itemize}

\subsection{Annotator Information}

All annotators are employed by a commercial data annotation company. We sign a contract with the company
and pay the company for the annotation work at a market price. The annotators are all college graduates
with strong English proficiency. A total of 15 annotators participated in the annotation work.

\section{Creation of \qanm}\label{movie_qa_pipeline}
The pipeline of StoryQA creation are:
\begin{itemize}
\item We prompt GPT-4 to generate as many diverse questions as possible and design distractors for each question.
\item We prompt GPT-4 to evaluate whether the ground-truth description provides enough information to answer each question. If it does not, the question-answer pair is discarded.
\item For each question, we add a new option, ``None of the Above". We then prompt GPT-4 to answer the question based on the ground-truth description. If GPT-4 selects ``None of the Above", the question-answer pair is discarded.
\item We prompt GPT-4 to answer the question directly without referring to any description. If it can provide the correct answer, the question-answer pair is discarded.
\item Finally, we employ GPT-4 to deduplicate similar questions.
\end{itemize}
We release the code and prompts used for the data generation process.

\section{Long Video Segmentation Method}\label{segmentation_method}

We segment long videos using two types of temporal cues: shot timestamps and the start and end times of character dialogues. The specific processing steps are as follows:
\begin{itemize}
\item Use the PySceneDetect\cite{scenedetect} to identify a series of cut points in the video.
\item For each cut point, verify if it occurs within the start and end time of a character dialogue. If so, discard that cut point to avoid splitting the dialogue.
\item Examine each segment obtained from the cut points. If a segment contains more than two dialogues, randomly select a new cut point between the end of one dialogue and the start of the next to ensure each segment contains at most two dialogues, thereby preventing overly long segments.
\end{itemize}
The code for the video segmentation module is available in our codebase.

\section{Global ID Generation}\label{global_id_genetation}

%To generate global ID for each subtitle, we first use an audio embedding model to generate embeddings for audio corresponding to each subtitle and assign the same global ID to audios with cosine similarity greater than a specific threshold. In the identification process, audios with the same ID will be mapped to the same character.

To generate global ID for each dialogue line in a long video, we first use an audio embedding model to create embeddings for each utterance. Subsequently, we perform clustering, assigning the same global ID to all lines whose embeddings belong to the same cluster, based on a cosine similarity exceeding a specified threshold.

\noindent{\bf Audio Embedding.} We fine-tune the ERes2NetV2 model \cite{chen2024eres2netv2} to obtain improved audio embeddings for our scenario. The training data is based on the \datasetnm training set, from which we construct a series of pairwise samples: positive pairs consist of two audio segments from the same person, while negative pairs consist of segments from different individuals. In total, we collect 246,834 data pairs, with 5\% randomly sampled as the development set, and the rest used for training. The ERes2NetV2 model is fine-tuned using CosineEmbeddingLoss as the loss function. We set the learning rate to 0.005 and the batch size to 32.

\noindent{\bf Global ID Generation.} To determine the optimal threshold for clustering, we utilize data from the \datasetnm development set to create a validation dataset for threshold selection. We pair all utterances, labeling pairs as positive if both utterances were spoken by the same character, and negative otherwise. We evaluate thresholds ranging from 0.5 to 1.0 on this dataset, with the results presented in Table~\ref{threshold_select}. A threshold of 0.85, which achieved the highest audio-visual character identification accuracy, was selected. In this context, high precision in identifying whether two utterances were spoken by the same person is crucial, while maintaining a balance with recall. The chosen threshold resulted in a precision of 0.900 and a recall of 0.408, satisfying these requirements.

Additionally, our audio embedding model outperforms all open-source models (CAM++\cite{wang2023cam++}, Ecapa-tdnn\cite{desplanques2020ecapa} and ERes2NetV2\cite{chen2024eres2netv2}) under this condition, as Table \ref{audio_embedding} shown.

\begin{table}[htbp]
\centering
\scalebox{1}{
\begin{tabular}{c|cccc}
\toprule[1pt]
\textbf{Threshold} & \textbf{Accuracy} & \textbf{Presion} & \textbf{Recall} & \textbf{F1} \\
\midrule[0.5pt]
0.50 & 0.553 & 0.765 & 0.746 & 0.756 \\
0.55 & 0.571 & 0.787 & 0.721 & 0.752 \\
0.60 & 0.592 & 0.806 & 0.692 & 0.745 \\
0.65 & 0.643 & 0.827 & 0.659 & 0.733 \\
0.70 & 0.672 & 0.847 & 0.617 & 0.714 \\
0.75 & 0.687 & 0.866 & 0.565 & 0.684 \\
0.80 & 0.696 & 0.882 & 0.497 & 0.636 \\
0.85 & \textbf{0.700} & 0.900 & 0.392 & 0.589 \\
0.90 & 0.693 & 0.920 & 0.287 & 0.437 \\
0.95 & 0.677 & 0.948 & 0.118 & 0.210 \\
\bottomrule[1pt]
\end{tabular}
}
\caption{The performance of clustering result for various thresholds tested on the validation dataset.}
\label{threshold_select}
\end{table}

\begin{table}[htbp]
\centering
\scalebox{1}{
\begin{tabular}{lcc}
\toprule[1pt]
\multicolumn{1}{c}{\textbf{Model}} & \multicolumn{1}{c}{\textbf{Recall@P=0.9}} & \multicolumn{1}{c}{\textbf{F1}} \\
\midrule[0.5pt]
% Resnet34\cite{he2016deep} &  &&\\
CAM++ & 0.22 & 0.36\\ % t=0.6
Ecapa-tdnn & 0.23 & 0.36\\ % t=0.57
ERes2NetV2  & 0.25 & 0.40 \\ % t=0.62
\midrule[0.5pt]
\textbf{Ours} & \textbf{0.41} & \textbf{0.56}  \\ % t=0.85
\bottomrule[1pt]
\end{tabular}
}
\caption{Clustering result with Precision=0.9 on test dataset.}
\label{audio_embedding}
\end{table}

\section{Training Details}\label{training_detail}

Our training process consists of three phases, as detailed below.

\noindent{\bf Pre-training the audio module.} In the pre-training stage, we use the same pre-training dataset of SALMONN~\cite{tang2023salmonn}, consisting of both 960-hour LibriSpeech training set~\cite{panayotov2015librispeech} and 1000-hour GigaSpeech M-set~\cite{chen2021gigaspeech} for speech recognition, as well as 2800-hour WavCaps~\cite{mei2024wavcaps}, AudioCaps~\cite{kim2019audiocaps} and Clotho~\cite{drossos2020clotho} for audio captioning. In this phase, we only pre-train the window-level Q-former, while freezing the audio encoders and LLM. For the window-level Q-former, we use only one trainable query, and set $L= 17$, which corresponds to approximately 0.33 seconds per window. This configuration outputs 88 textual tokens from the Q-former for a 30-second audio segment. The model was trained on 16 H800 GPUs for 50,000 steps using a batch size of 6. 

\noindent{\bf Fine-tuning on audio and visual tasks.} To enhance the model's ability in audio and visual character identification, we collect training data for the following tasks: audio diarization, and character recognition, identification, and tracking. For audio diarization, we sampled data from AMI~\cite{ami2005}, AISHELL-4~\cite{fu2021aishell}, AliMeeting~\cite{Yu2022M2MeT,Yu2022Summary}, CallHome~\cite{callhome}, DIHARD~\cite{ryant2019second}, VoxConverse~\cite{chung2020spot}, and LibriCSS~\cite{chen2020continuous}, creating a training dataset consisting of 290,088 minute-level utterances. Additionally, using MovieNet~\cite{huang2020movienet}, we construct training data for three tasks: (1) character recognition, which involves recognizing characters from frames; (2) character identification, where given a cast list, the model identifies the characters in the frames; and (3) character tracking, where the model tracks a character's position across all frames based on a given face. We collect 33,576 data for these three tasks.

We jointly trained on all four tasks by sampling training data from each task in equal proportions for each batch. During training, we fine-tuned the visual adapter, audio window-level Q-former, and the LLM. The model was trained on 16 H800 GPUs for 8,000 steps using a batch size of 6.

\noindent{\bf Fine-tuning for audio-visual character identification.} In this stage for the audio-visual character identification task, we utilize audio-visual character identification data from the \datasetnm training set, including 5,350 videos. During this stage, we train adapters for the visual and audio modules as well as the LLM. The fine-tuning was performed using 16 H800 GPUs for 4000 steps with a batch size of 4.

\section{Prompt Templates}\label{templates}

\begin{table*}[b]
\centering
\resizebox{\linewidth}{!}{%
\begin{tabular}{p{19cm}}
\toprule[1pt]
\textbf{The prompt for Gemini-1.5-pro.} (The input `video\_clip' is a minutes-long video.)\\
\midrule[0.5pt]
[Instruction]
This video is a clip from a film. Please provide a detailed plot description of this clip in time slots, with each time slot being as short as possible to ensure that the described plot is sufficiently detailed. The plot description can include but is not limited to:

- Names, identities, and relationships of the characters

- Interactions between characters and the scene, events occurring, actions and dialogues between characters

- Psychological states and emotional changes of the characters

- Changes in the plot, scene transitions, and storyline progression

- When describing the plot, please include actions, emotions, scenes, and other aspects, not just the subtitle information

Output in the following format without any extra information:

mm:ss~mm:ss Event1

mm:ss~mm:ss Event2

mm:ss~mm:ss Event3

...

\{video\_clip\}

[Cast List]

Below is the cast list for this film. When describing the plot, please refer to the appearance of the actors and their character names. Note that not all characters in the cast list will necessarily appear in this scene, and for characters in the scene who are not listed in the cast list, you may use other aliases.

Character name: \{character\_name\}

\{character\_photo\}

...

[Subtitle]

Here is the text transcription of the audio from the video:

\{subtitles\}
\\
\midrule[1pt]
\textbf{The prompt for GPT-4o.}  (The input `video\_clip' is a seconds-long video clip after segmentation.)\\
\midrule[0.5pt]
Given a video and supplementary information, please finish the following tasks.

Video: \{video\_clip\}

Known characters: \{character\_photo\} \{character\_name\}.

Subtitles of the input video with time stamps:

\{subtitles\}

Tasks:

1. Describe the events that occur within this video.\\
\bottomrule[1pt]
\end{tabular}
}
\caption{The prompt for description generation for open source models.}
\label{description_prompts}
\end{table*}

\begin{table*}[b]
\centering
\resizebox{\linewidth}{!}{%
\begin{tabular}{p{19cm}}
\toprule[1pt]
\textbf{The prompt for GPT-4 evaluation.}\\
\midrule[0.5pt]
[Movie Plot Description]

\{text\}

[Multiple-Choice Question]

\{question\}

Using the information in the [Movie Plot Description], please answer the [Mulitple-Choice Question]. Only one of options (A, B, C, D) is correct. Your response should follow this format:

[Reason]: Explain your reasoning.

[Answer]: Generate only one character from the options (A, B, C, D).\\
\bottomrule[1pt]
\end{tabular}
}
\caption{The promtp for GPT-4 to answer the questions related to the video based on the generated descriptions.}
\label{evaluation_prompts}
\end{table*}

Table~\ref{description_prompts} shows the prompt templates for Gemini-1.5-pro and GPT-4o. Gemini takes the entire 3-minute video in the \datasetnm as input, but due to the frame rate limitation of the interface, we segment the long video into seconds-long clips, extract 8 frames from each short video and have GPT-4o generate descriptions separately.

Table~\ref{evaluation_prompts} shows the prompt templates for GPT-4 to answer the questuions in \datasetnm. GPT-4 takes the  whole descriptions and a multiple-choice question as input and output the selected answer and reason.

\end{document}